\documentclass[letterpaper, 10 pt, conference]{ieeeconf}
\usepackage{amssymb}
\usepackage{amsmath}
\usepackage{amssymb}
\usepackage{mathtools}
\usepackage{graphicx}
\usepackage{algorithm}
\usepackage{paralist}
\let\labelindent\relax
\usepackage{enumitem}
\usepackage[noend]{algorithmic}

\usepackage{microtype}
\usepackage{subfigure}
\usepackage{booktabs}

\usepackage{hyperref}

\usepackage[textsize=tiny]{todonotes}
\usepackage{xspace}
\newcommand{\method}{\text{TADPO\xspace}}

\usepackage{pgfplots}
\usepackage{subcaption}
\usepackage{multirow}
\usepackage{wrapfig}
\usepackage{makecell}

\IEEEoverridecommandlockouts                              % This command is only needed if
\title{\LARGE \bf
TADPO: Reinforcement Learning Goes Off-road
}

\author{
Zhouchonghao Wu$^{*1}$, Raymond Song$^{*1}$, Vedant Mundheda$^{*1}$, Luis E. Navarro-Serment$^{1,2}$,\\Christof Schoenborn$^{1,2}$, and Jeff Schneider$^{1}$% <-this % stops a space
\thanks{*: Equal contributions, order randomized.}% <-this % stops a space
\thanks{$^{1}$Robotics Institute, School of Computer Science, Carnegie Mellon University, Pittsburgh, PA 15213.
       {\tt\small \{zhouchow, rysong, vmundhed, jeff4\}@andrew.cmu.edu}}%
\thanks{$^{2}$National Robotics Engineering Center, Robotics Institute, School of Computer Science, Carnegie Mellon University, Pittsburgh, PA 15201.
        {\tt\small \{lenscmu, cschoenb\}@nrec.ri.cmu.edu}}%
}
\begin{document}

\maketitle
\thispagestyle{empty}
\pagestyle{empty}

%%%%%%%%%%%%%%%%%%%%%%%%%%%%%%%%%%%%%%%%%%%%%%%%%%%%%%%%%%%%%%%%%%%%%%%%%%%%%%%%
\begin{abstract}

Off-road autonomous driving poses significant challenges such as navigating unmapped, variable terrain with uncertain and diverse dynamics.
Addressing these challenges requires effective long-horizon planning and adaptable control.
Reinforcement Learning (RL) offers a promising solution by learning control policies directly from interaction.
However, because off-road driving is a long-horizon task with low-signal rewards, standard RL methods are challenging to apply in this setting.
We introduce \method{}, a novel policy gradient formulation that extends Proximal Policy Optimization (PPO), leveraging off-policy trajectories for teacher guidance and on-policy trajectories for student exploration.
Building on this, we develop a vision-based, end-to-end RL system for high-speed off-road driving, capable of navigating extreme slopes and obstacle-rich terrain.
We demonstrate our performance in simulation and, importantly, zero-shot sim-to-real transfer on a full-scale off-road vehicle. To our knowledge, this work represents the first deployment of RL-based policies on a full-scale off-road platform.
Source code is available at this \href{https://github.com/tadpo-algorithm/tadpo}{link}
and video at this \href{https://youtu.be/I54T--_PXYM}{link}.

\end{abstract}
%%%%%%%%%%%%%%%%%%%%%%%%%%%%%%%%%%%%%%%%%%%%%%%%%%%%%%%%%%%%%%%%%%%%%%%%%%%%%%%%
\section{Introduction}

Autonomous ground vehicles have achieved remarkable progress in structured environments such as highways and urban roads, with detailed maps, high-quality annotations, and where the vehicle-terrain dynamics are easy to model.
In contrast, off-road autonomy remains an open challenge. Vehicles must navigate unstructured environments such as sand, gravel, vegetation, and steep slopes, where terrain–vehicle interactions are complex, uncertain, and difficult to model.
These conditions require both adaptive control strategies and long-horizon planning without relying on the dense mapping and annotation pipelines available in urban driving.
Safe navigation in these settings depends on the vehicle's ability to perceive and reason about traversable regions in real time, while avoiding obstacles at high speeds.

These challenges motivate the use of Reinforcement Learning (RL), which can directly learn control policies from interaction, bypassing the need for explicit dynamics models, dense maps, or costly labeling. RL offers the potential to leverage large-scale simulation data for training while generalizing to real-world conditions at deployment.

At the same time, applying RL to off-road autonomy is challenging due to low-signal rewards, long-horizon decision-making, complex terrain dynamics, and decision making in unstructured environments. Such conditions exacerbate difficulties in exploration, and standard RL methods often fail to acquire robust policies without additional guidance.

\begin{figure}[t]
    \centering
    \includegraphics[width=1.0\linewidth]{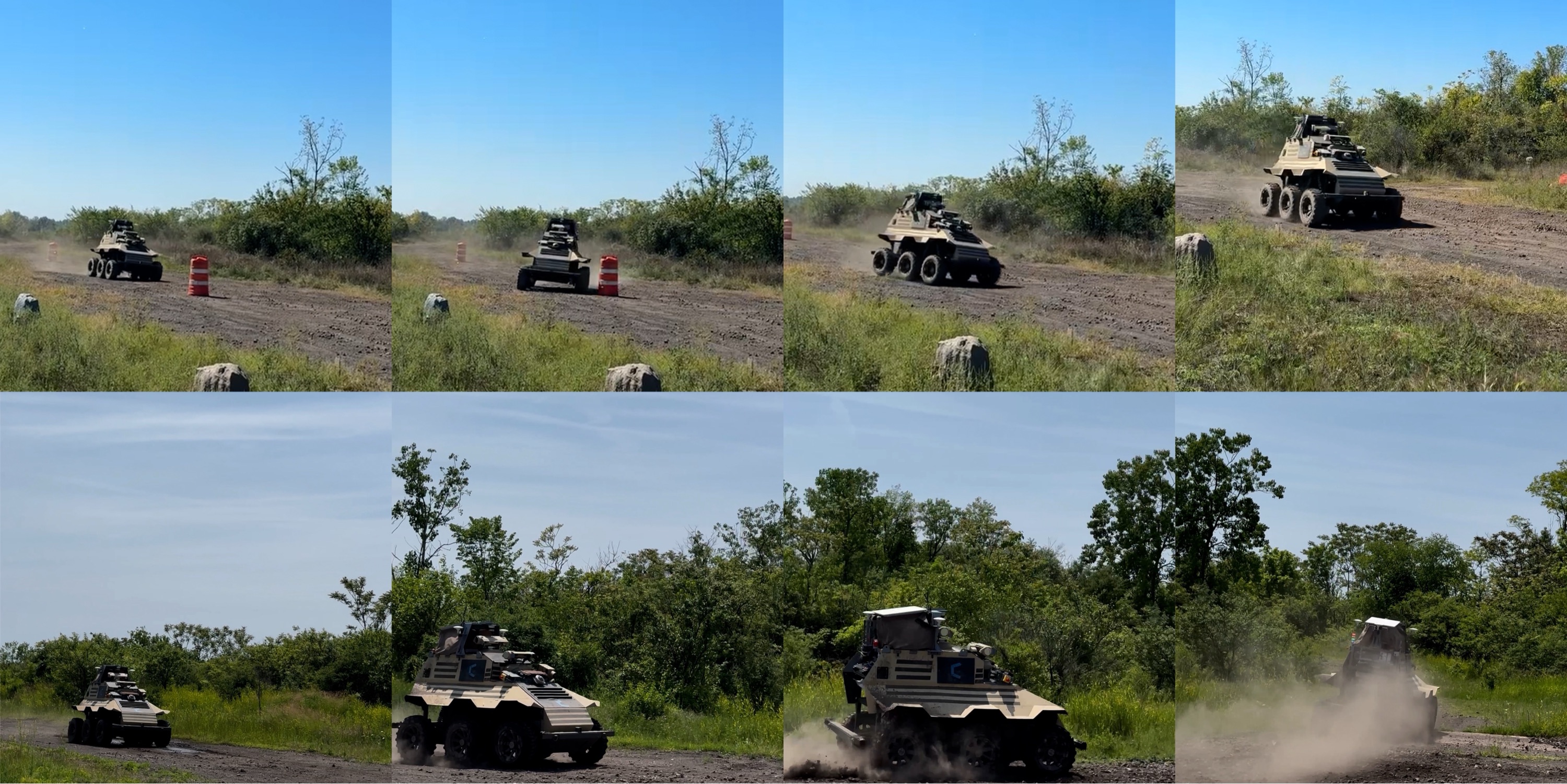}
    \caption{
    Autonomous vehicle avoiding obstacles (top) and taking corners at speed (bottom) controlled using TADPO-trained end-to-end policies.
    }
    \label{fig:sabercat}
    \vspace{-5mm}
\end{figure}

A promising strategy is teacher-guided RL, where demonstrations or expert actions are distilled into a student policy while continuing to explore beyond the teacher's demonstrations.
This combination enables RL policies to benefit from expert guidance during training while operating without privileged information.
Crucially, we find that such a framework also enables strong sim-to-real transfer, allowing policies trained entirely in simulation to be deployed on real full-scale off-road vehicles without fine-tuning.
Fig \ref{fig:sabercat} demonstrates our vehicle avoiding obstacles and doing long-distance high-speed control using a policy trained by our method \method{}.

Following this approach, we present three major contributions in this paper:
\begin{itemize}[leftmargin=1em, itemsep=0pt, topsep=0pt]
    \item Teacher Action Distillation with Policy Optimization (\method), a novel extension of Proximal Policy Optimization (PPO) that enables concurrent learning from fixed demonstrations and on-policy interactions to tackle long-horizon planning and hard exploration problems.
    \item A vision-based, end-to-end RL system for high-speed off-road driving.
    We demonstrate high performance navigation through extreme slopes and obstacle-rich terrains in simulation.
    \item The first successful deployment, to the author's knowledge, on a full-scale off-road vehicle of RL-based policies, demonstrating end-to-end and zero-shot sim-to-real capabilities.
\end{itemize}

\section{Related Work}
\paragraph*{Off-Road Driving}
Multiple works have explored end-to-end RL methods for off-road \cite{rl_offroad:wroom, imitation_offroad:human_demonstrations, rl_offroad:unrealistic} and on-road \cite{learningdriveday, onroad:merge_highway, onroad_rl:urban} driving.
Off-road works are limited and often focus on immediate obstacle avoidance, lacking long-term planning, or are tested in unrealistic simulations.
Meanwhile, on-road driving research focuses on the unpredictable behavior of other road users, rather than the variability of terrain encountered in off-road environments.
An issue among RL methods presented in these works is their inability to explore efficiently, rendering them ineffective in obstacle-rich environments where simulation is computationally expensive and dynamics are highly complex.
As a result, exploration in these scenarios is challenging without external guidance.
Another issue common among RL methods is that autonomous driving is often goal-conditioned, requiring longer-horizon planning and reasoning for successful off-road navigation.
Some existing works \cite{nasiriany2019planninggoalconditionedpolicies, chanesane2021goalconditionedreinforcementlearningimagined} seek to tackle this issue, but in end-to-end navigation tasks that involve image inputs, such methods require sampling image inputs becoming computationally intractable.

Sampling methods like Model Predictive Path Integral (MPPI) \cite{algo:mppi} and Cross-Entropy Method (CEM) \cite{algo:cem} have been applied in the off-road autonomy domain in previous works.
MPPI, in particular, has been applied in various domains like UAVs \cite{mpc_uav,mpc_aerialmanipulator} and the off-road autonomy domain \cite{mpc_offroad:aggressive_uw, mpc_offroad:lee_high_speed}.
Long Range Navigator (LRN) \cite{offroad:longrange} learns affordance-based intermediate representations from unlabeled ego-centric videos to guide long-horizon off-road planning beyond local metric maps, reducing myopic decisions and improving navigation efficiency.
Although the sampling methods vary between these techniques, they all require sampling a large number of trajectories to select a feasible action sequence that minimizes a cost function.
While effective for generating control actions in complex, nonlinear systems, the dense sampling they require makes high-quality, real-time operation for long-horizon planning computationally impractical.
Some attempts like RL+MPPI \cite{mppi:rl+mppi} and TD-MPC \cite{mppi:tdmpc} have been made to improve sampling efficiency by learning a state-dependent control action distribution and learning a terminal value function, thereby reducing the required number of samples and planning horizon.

There are also recent works that learn costmaps instead of policies directly from experience. Self-supervised methods predict terrain costs from exteroceptive and proprioceptive signals \cite{costmap:HDIF}, while SALON adapts online to new environments using speedmaps and uncertainty avoidance \cite{costmap:SALON}. Risk-aware IRL further refines cost functions by accounting for uncertainty and safety \cite{costmap:IRL}.
RoadRunner \cite{cv_offroad:traversability1} predicts terrain traversability and elevation maps from camera and LiDAR inputs in a self-supervised, low-latency pipeline. In parallel, Zhu et al. \cite{cv_offroad:traversability2s} use Deep Inverse Reinforcement Learning to learn traversability cost functions from expert demonstrations, incorporating vehicle kinematics for trajectory planning.

\paragraph*{Reinforcement Learning}
Proximal Policy Optimization (PPO) and Soft Actor-Critic (SAC) are common modern RL techniques for solving complex robotics tasks.
PPO \cite{algo:ppo} is an on-policy RL framework that performs stable policy learning by limiting policy updates through a clipped surrogate objective function.
SAC \cite{algo:sac} is an off-policy RL algorithm that optimizes a stochastic policy and value function, enabling efficient and stable learning for continuous control tasks.
Both methods have been successfully applied to robotics tasks involving visual inputs and continuous action spaces, including dexterous manipulators, bipedal robots, and unmanned aerial and ground vehicles \cite{ppo:manipulation, ppo:bipedal, ppo:uav, ppo:ground}.

Several works augment RL with external guidance through demonstrations or teacher-student frameworks \cite{tsrl:cts,tsrl:ppd}. Approaches such as DAgger \cite{algo:dagger}, offline RL methods like IQL \cite{algo:iql}, and PPO variants with demonstrations \cite{ppo_single_demo} highlight the benefits of combining imitation and reinforcement learning. Despite their successes, these algorithms encounter unique challenges in the proposed off-road driving problem, including navigating diverse terrains and long-horizon planning \cite{safe_driving_egpo,dqfd,policyopt_fromdemos,sac_lfd,ppovssac}.

Several works use Visual Foundation models such as DinoV2 \cite{vfm:dinov2}, SAM \cite{vfm:sam}, and SAM2 \cite{vfm:sam2} to extract image features for vision-based RL.
These foundational models help bridge the domain gap between simulation and real-world.

%===============================================================================

\begin{figure*}[h!]
    \centering
    \vspace{3mm}
    \includegraphics[width=0.94\linewidth]{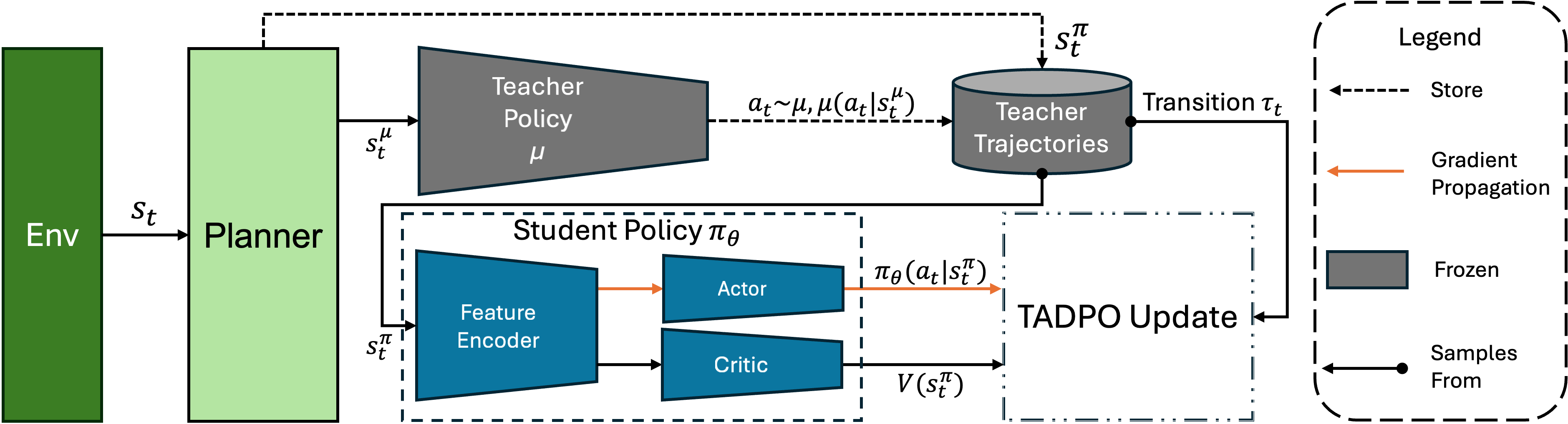}
    \caption{
   \textbf{ Teacher Action Distillation Rollout and Update Process.}
    The teacher demonstration buffer is frozen while training the student policy.
    The student policy performs a TADPO update with a probability $p$ solely on the actor and the feature encoder of the policy, using the critic to estimate the advantage of the teacher rollout over the student for any environment state.
    }
    \label{fig:tadpo_update}
    \vspace{-7mm}
\end{figure*}

\section{Background}
\label{ppo_explain}

We model the control problem of an off-road autonomous vehicle as a Markov Decision Process (MDP), represented by the tuple \( \mathcal{M} = (\mathcal{S}, \mathcal{A}, P, r, \gamma) \), where
\( \mathcal{S} \) is the state space,
\( \mathcal{A} \) is the action space,
\( P(s' | s, a) \) is the transition dynamics function,
\( r: \mathcal{S} \times \mathcal{A} \rightarrow \mathbb{R} \) is the reward function,
and
\( \gamma \in [0, 1) \) is the discount factor.

Proximal Policy Optimization (PPO) \cite{algo:ppo} is an on-policy algorithm in the Policy Gradient family. Given a $\theta$-parameterized policy $\pi_\theta$ and a set of trajectories collected by it, PPO employs an actor-critic architecture where the actor learns the policy and the critic estimates the value function to guide policy improvement, maximizing a clipped surrogate objective in Equation \eqref{eq:ppoloss} to update the policy.

\resizebox{.9\linewidth}{!}{
  \begin{minipage}{\linewidth}
\begin{align}
    L^{\text{CLIP}}(\theta) &= \mathbb{E}_t \left[ \min \left( r_t(\theta) \hat{A}_t, \text{clip} (r_t(\theta), 1 - \epsilon, 1 + \epsilon) \hat{A}_t \right) \right] \label{eq:lppo} \\
    L^{\text{VF}}(\theta) &= \mathbb{E}_t \left[ (V_{\pi_{\theta_\text{old}}}(s_t) - R_t)^2 \right] \label{eq:lvf} \\
    L^{\text{entropy}}(\theta) &= \mathbb{E}_t \left[ - H[\pi_\theta(\cdot | s_t)] \right] \label{eq:lentropy} \\
    L^{\text{PPO}}(\theta) &= L^{\text{CLIP}}(\theta) - c_1 L^{\text{VF}}(\theta) + c_2 L^{\text{entropy}}(\theta) \label{eq:ppoloss}
\end{align}
\end{minipage}
}

where $r_t(\theta) = \frac{\pi_\theta(a_t | s_t)}{\pi_{\theta_{\text{old}}}(a_t | s_t)}$ is the ratio of the probability of the action $a_t$ under the current policy ($\pi_\theta$) to the policy used to collect the rollout ($\pi_{\theta_{\text{old}}}$),
$\hat{A}_t=\sum_{i=t}^{t+T} (\gamma \lambda)^{i-t} \delta_{i}$ is the estimate of the advantage, with $\delta_t = R_t + \gamma V_{\pi_{\theta_{\text{old}}}}(s_{t+1}) - V_{\pi_{\theta_{\text{old}}}}(s_t)$,
$R_t = \sum_{i=t}^{t+T} \gamma^{i-t} r(s_i, a_i) + \gamma^{T-t+1}V(s_{T+1})$ is the discounted return,
and $T$ is the number of transitions,
$H[\pi_\theta(\cdot | s_t)]$ is the entropy of the action distribution induced by the policy given the state \( s_t \), $V_{\pi_{\theta_\text{old}}}(s_t)$ is the expected return of the policy at state $s_t$, and $c_1, c_2$ are constants.

The advantage estimator $\hat{A}_t$ quantifies the benefit of executing action $a_t$ over the expected return $V_{\pi_{\theta_\text{old}}}(s_t)$ of $\pi_{\theta_\text{old}}$ at $s_t$.
This implies that the policy gradient update derived from Equation \eqref{eq:ppoloss} remains sound only when  $V_{\pi_{\theta_\text{old}}}$ maintains a faithful approximation of the policy's expected returns, a key motivation behind our formulation in Equation \eqref{eq:reladv}.

PPO struggles with exploration in complex long-horizon tasks, often failing to learn effective policies \cite{ppo_single_demo}.
Its undirected randomness leads to inefficient sampling, with actions rarely aligned to task objectives.
While expert knowledge could help, PPO's on-policy nature prevents leveraging off-policy data like demonstrations, limiting its use in domains where effective exploration is challenging.

\section{\method: Teacher Action Distillation with Policy Optimization}

We propose a novel method to train a $\theta$-parameterized policy $\pi_\theta$ by extending PPO to incorporate demonstrations from a teacher policy $\mu$. This extension allows $\pi_\theta$ to be trained concurrently using its own on-policy rollouts and demonstrations by the teacher policy.

\subsection{Teacher Action Distillation Policy Gradient}

Given a pre-trained teacher policy $\mu$, we define a loss function $L^{\text{TAD}}$ in Equation \eqref{eq:ltadpo_tot} to train a student policy $\pi_\theta$. This loss is computed solely on teacher rollouts, where actions at each time step $t$ are sampled from the teacher policy, i.e., $a_t \sim \mu$.

\begin{align}
    L^{\text{TAD}}(\theta) &= L^{\mu}(\theta) + c_2 L^{\text{entropy}}(\theta) \label{eq:ltadpo_tot} \\
    \rho_t(\theta) &= \frac{\pi_\theta(a_t | s_t^\pi)}{\mu(a_t | s_t^\mu)} \label{eq:rho}\\
    \hat{\Delta}_t &= R(a_t, s_t) - V_{\pi_{\theta_\text{old}}} (s_t^\pi) \label{eq:reladv}\\
    L^{\mu}(\theta) &= \mathbb{E}_{a_t \sim \mu} \left[ \max \left( 0, \min (\rho_t(\theta),1 + \epsilon_\mu) \hat{\Delta}_t \right) \right] \label{eq:lmu}
\end{align}

where $L^{\text{entropy}}$ is defined in Equation \eqref{eq:lentropy} and $\epsilon_\mu$ is a hyperparameter.
It is important to note that $\mu$ and $\pi_\theta$ can be defined to operate on distinct observation spaces, denoted by $s^\mu_t$ and $s^\pi_t$ respectively, despite being derived from the same underlying environment state $s_t$.
The relaxed assumptions on the relative structure of the teacher and student models allow the teacher to utilize privileged observations or higher-capacity architectures.

In Equation \eqref{eq:rho}, $\rho_t(\theta)$ is defined as the ratio of probability of $a_t$ under $\pi_\theta$ to the probability under $\mu$.
This concept is analogous to the probability ratio $r_t$ in PPO.
In Equation \eqref{eq:reladv}, $\hat{\Delta}$ estimates the advantage of the achieved return of the teacher rollout at state $s_t$ over the expected return of the student.
We observe that applying the policy gradient update from $L_\mu$ to the student policy exclusively when $\hat{\Delta} > 0$ (i.e., when the teacher outperforms the student's expectations) results in stable updates and higher reward attainment by the student.

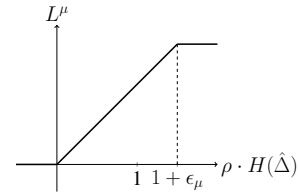
\begin{figure}[h]
    \centering
    \resizebox{0.45\linewidth}{!}{  % Match the wrap width
        \begin{tikzpicture}
            \begin{axis}[
                every axis/.append style={font=\huge},
                axis lines=middle,
                xlabel={$\rho \cdot H(\hat{\Delta})$},
                ylabel={$L^{\mu}$},
                xtick={0, 1, 1.5},
                xticklabels={0, 1, $1 + \epsilon_\mu$},
                ytick=\empty,
                ymin=-0.1, ymax=1.5,
                xmin=-0.5, xmax=2,
                samples=100,
                domain=-1.5:2,
                thick,
                axis line style={->},
                xlabel style={at={(ticklabel* cs:1)}, anchor=west},
                ylabel style={at={(ticklabel* cs:1)}, anchor=south},
                axis equal,
            ]
                \addplot[domain=-1.5:0, line width=1.5pt] {0};
                \addplot[domain=0:1.5, line width=1.5pt] {x};
                \addplot[domain=1.5:2, line width=1.5pt] {1.5};
                \addplot[dashed] coordinates {(1.5,0) (1.5,1.5)};
            \end{axis}
        \end{tikzpicture}
    }
    \caption{\small A single timestep of the teacher distillation loss function $L^\mu$ as a function of $\rho \cdot H(\hat{\Delta})$, where $H(\cdot)$ is the Heaviside step function.}
    \label{fig:tadpo_loss}
\end{figure}
In Equation \eqref{eq:lmu}, $L_\mu$ is clipped as illustrated in Figure \ref{fig:tadpo_loss}. This clipping mechanism prevents policy gradient updates when $\rho_t$ exceeds $1 + \epsilon_\mu$, effectively halting updates when the probability ratio of action $a_t$ under $\pi_\theta$ relative to $\mu$ surpasses this threshold, indicating that the student policy has already captured the desired behavior.

Thus, $L^{\text{TAD}}$ in Equation \eqref{eq:ltadpo_tot} ensures that the policy gradient propagates only when two conditions are met:
(i) the teacher rollout's return exceeds the student policy's expected return, and
(ii) the student policy's probability of executing action $a_t$ is not substantially higher than that of the teacher policy.
Analogous to PPO, $L^\text{entropy}$ in Equation \eqref{eq:ltadpo_tot} modulates the student policy's exploration.

\begin{algorithm}
\caption{TADPO \label{alg:tadpo}}
\resizebox{0.9\textwidth}{!}{%
\begin{minipage}{\textwidth}
\begin{algorithmic}[0]
\STATE \textbf{Input:} Teacher $\mu$, Student $\pi$, Sampling prob. $p$
\STATE \textbf{Return:} Student policy params $\theta$
\STATE $\mathcal{B}_\mu \leftarrow$ $N_\mu$ transitions: $\{\tau_{t_{a_t \sim \mu}} = (s^\mu_t, a_t, R_t, \mu(a_t | s^\mu_t))\}$
\FOR{iter $= 1$ to $I$}
    \STATE $\mathcal{B}_\pi \leftarrow$ $N_\pi$ transitions:\\
    $\{\tau_{t_{a_t \sim \pi_{\theta_\text{old}}}}=(s^\pi_t, a_t, R_t, \pi_{\theta_\text{old}}(a_t | s^\pi_t)) \}$
    \FOR{epoch $= 1$ to $K$}
        \WHILE{$\mathcal{B}_\pi \neq \emptyset$}
            \STATE Sample $r \sim \mathcal{U}(0,1)$
            \IF{$r > p$}
                \STATE Sample $n$ transitions $\tau \sim \mathcal{B}_\pi$ w/o replacement
                \STATE $\theta \leftarrow \text{PPOUpdate}(\tau)$
            \ELSE
                \STATE Sample $n$ transitions $\tau \sim \mathcal{B}_\mu$ w/o replacement
                \STATE $\theta \leftarrow \text{TADPOUpdate}(\tau)$
            \ENDIF
        \ENDWHILE
        \STATE Reset $\mathcal{B}_\mu$, $\mathcal{B}_\pi$
    \ENDFOR
\ENDFOR
\end{algorithmic}
\end{minipage}
}

\end{algorithm}

\subsection{Training Procedure}
\label{training_procedure}

\method{} involves learning a student policy $\pi_\theta$ through concurrent utilization of teacher demonstrations and student rollouts. Transitions are collected from both the pre-trained teacher policy $\mu$ and the student policy $\pi_\theta$ being trained, and stored in separate teacher and student buffers, respectively. With probability $p$, the algorithm samples transitions from the teacher's buffer and performs a TADPO update, which likely involves distilling knowledge from the teacher. Otherwise, it samples from the student's buffer and performs a standard PPO update. As shown in Algorithm \ref{alg:tadpo}, this alternating process continues for multiple iterations and epochs, allowing the student to learn from both its own experiences and the teacher's expertise. In our implementation, $\hat{\Delta}_t$ is normalized to have unit standard deviation within each mini-batch.
As shown in Figure \ref{fig:tadpo_update}, during the TADPO update, gradient propagation occurs exclusively through the actor and feature encoder components of the student policy $\pi_\theta$, while the critic remains frozen, ensuring the value function maintains independent state-value estimates based solely on the student's experiences.

\section{End-to-end Off-road Autonomy}

We employ a hierarchical architecture to achieve end-to-end off-road autonomy. Given a final goal $\textbf{p}_g$, a global planner generates sparse waypoints utilizing a coarse global map.
These waypoints are tracked by an RL controller trained using TADPO. As the globally planned sparse waypoints may be suboptimal and fail to account for all obstacles, the RL controller must incorporate long-horizon planning capabilities to effectively track these waypoints.

\begin{figure}[t]
    \vspace{2mm}
    \centering
    \includegraphics[width=0.8\linewidth]{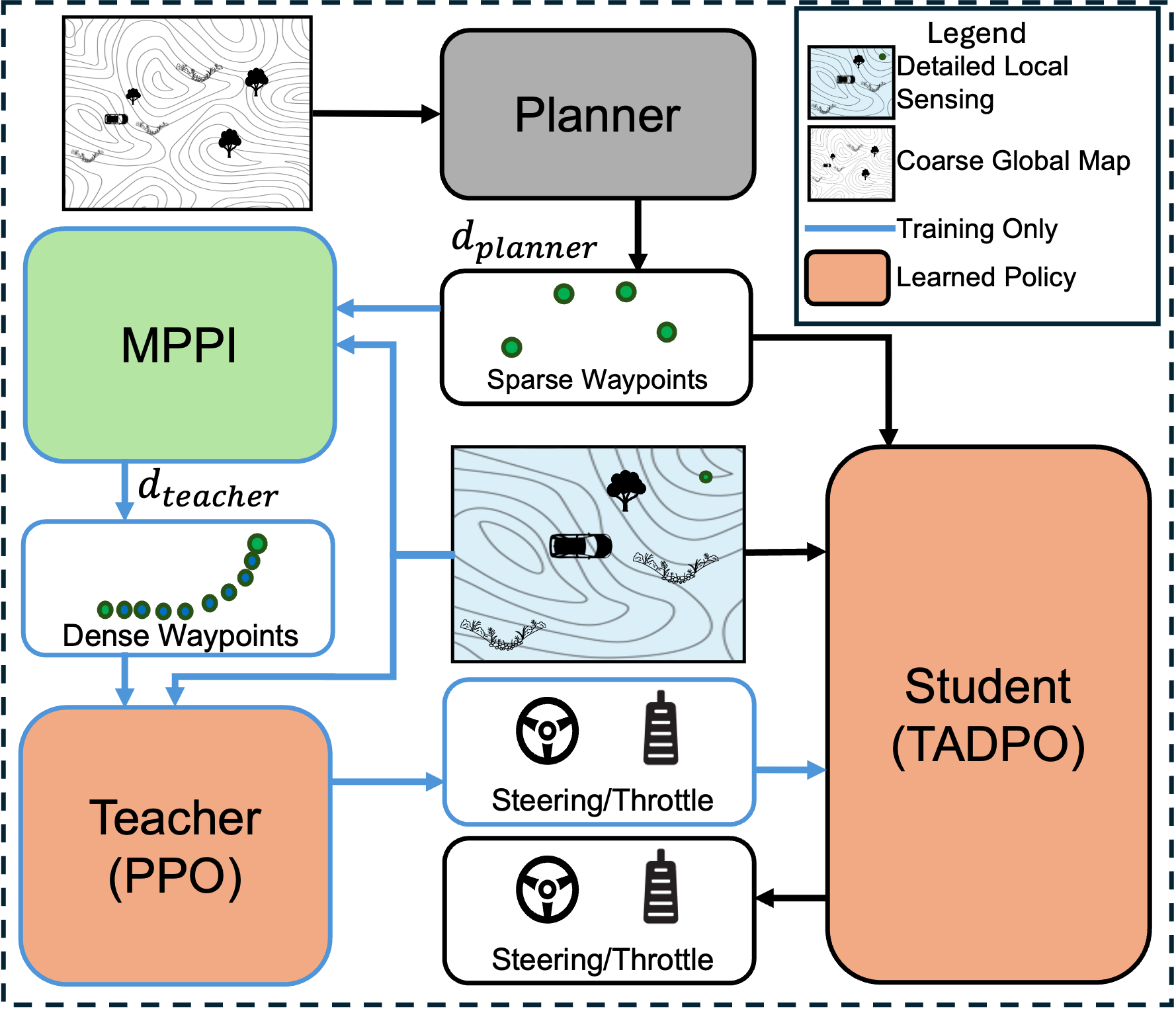}
    \caption{
    Hierarchical Autonomy Pipeline.
    During training, MPPI generates dense waypoints for a teacher policy to follow, providing demonstrations for TADPO, which tracks sparse waypoints.
    During deployment, TADPO tracks sparse waypoints directly without MPPI.
    In simulation, $d_{planner}=80$ and $d_{teacher}=6$. In real-world deployment, $d_{planner}=20$ and $d_{teacher}=4$
    }
    \label{fig:sim}
    \vspace{-2mm}
\end{figure}

\begin{figure}[t]
    \centering
    \includegraphics[width=0.8\linewidth]{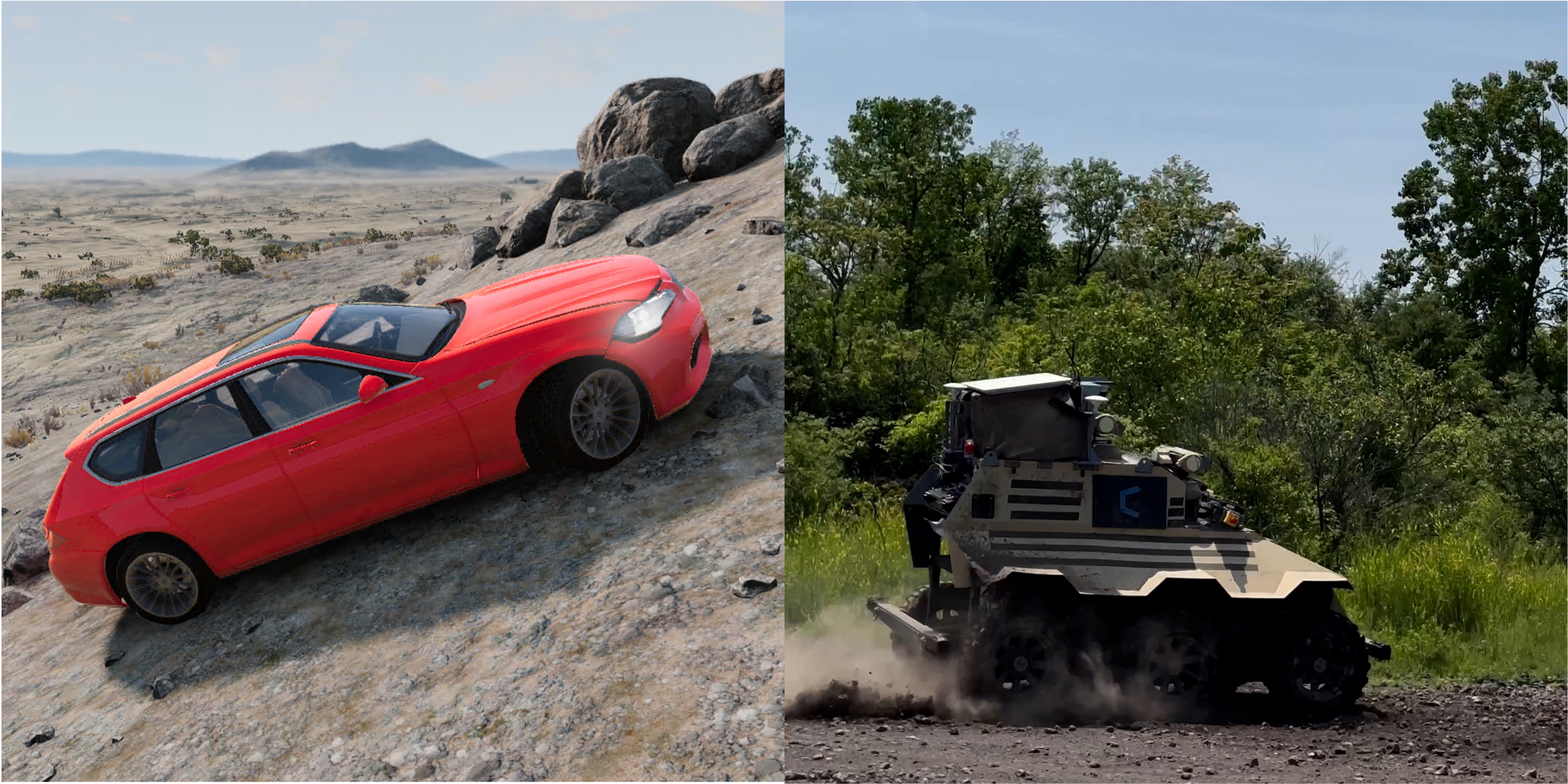}
    \caption{
    A comparison of the training vehicle in simulation environment and the deployment vehicle in deployed environment. A large embodiment gap can be observed both vehicle dynamics and the terrains.
    }
    \label{fig:sim-real}
    \vspace{-5mm}
\end{figure}

\subsection{Training}
\label{offroad_training}

An MPPI controller interpolates sparse waypoints using the cost function from \cite{mpc_offroad:aggressive_uw} to generate dense waypoints for training the teacher policy.
The PPO teacher policy $\mu$ is trained using dense MPPI waypoints.
The student policy $\pi_\theta$ is then trained to distill the teacher behavior via the TADPO training procedure in Section \ref{training_procedure} while operating solely with sparse waypoints provided by the global planner.
Hence, training with sparse waypoints, requires $\pi_\theta$ to learn sophisticated planning capabilities to maneuver through ditches and avoid obstacles.
Figure \ref{fig:sim} offers an illustration of how the TADPO training pipeline works in the off-road driving task.

\subsection{Reward Function, Observation and Action Spaces, and Model Architecture}
\label{text:reward}
The reward function includes progress toward waypoints, penalties for collisions, damage, and jerk, and a success bonus.
Observations combine proprioceptive and visual inputs. Proprioception includes normalized speed, roll, pitch, and waypoint encodings, with teachers using dense and students using sparse waypoint plans.
Visual inputs include a stack of three frames from top-down and forward views: teachers get high-resolution local maps; students use wider, lower-resolution ones.
We use NatureCNN \cite{rl:atari} as our feature encoder in simulation.
The controller outputs throttle and steering commands.
The specific hyperparameters are described in Table \ref{tab:ppo_hyperparameters}.

\begin{table}
    \vspace{2mm}
    \centering
    \begin{tabular}{lc}
    \toprule
    {Hyperparameters} & Value  \\
    \midrule
        Update ratio ($\epsilon_\mu$) & 0.5 \\
        Teacher policy ratio ($p$) & 0.5 \\
        Learning Rate & 3e-4 \\
        Discount Factor ($\gamma$) & 0.99 \\
        Clip Range ($\epsilon$) & 0.2 \\
        Number of Epochs & 20 \\
        Mini-batch Size & 256 \\
        Number of Steps per Update & 2048 \\
        Value Function Coefficient ($\lambda_v$) & 0.5 \\
        Entropy Coefficient ($\lambda_e$) & 0.001 \\
        Teacher Demonstration Buffer Size & 1e5 \\
    \midrule
        CNN Vision Backbone & \makecell[c]{\texttt{NatureCNN} with \\ 256 dim latent space \cite{rl:atari}} \\
        MLP Head Architecture & [128,64,64] \\
    \midrule
        Vision-backbone & \texttt{DinoV2-ViT-S/14} (frozen) \\
        Decoder MLP & [768, 4] \\
  \bottomrule
    \end{tabular}
    \caption{
    Hyperparameters for Teacher and Student Training. Common hyperparameters between simulation training and real-world training are in section 1.
    Network architecture for simulation and real-world policies are listed in section 2 and 3 respectively.
    }
    \label{tab:ppo_hyperparameters}
    \vspace{-3mm}
\end{table}

\section{Training and Evaluation in Simulation}

\subsection{Simulator}
\label{sim_traj}

We use BeamNG.tech \cite{sw:beamng} as the simulator for training and evaluating our algorithms. BeamNG.tech offers a highly realistic simulation environment, featuring advanced vehicle dynamics, sensor simulation, and damage modeling, allowing us to train and test our algorithms in a realistic environment that closely mirrors real-world conditions.
We use \verb|etk800| as our vehicle in Simulation. A comparison of the simulation vehicle and the deployment vehicle is shown in Figure \ref{fig:sim-real}.

\subsection{Training, Demonstration, and Testing Datasets}
\label{text:dataset}

We train teacher and student policies in a simulated desert environment. Sparse waypoints ($80$ m apart) are generated using an A* planner over a coarse global map. A fixed set of start-goal pairs is used for teacher training and demonstration trajectories. Dense waypoints ($6$ m apart) between sparse ones are generated via an MPPI controller using semantic segmentation and depth. A separate trajectory set is used for evaluation. Expert demonstrations use these MPPI-generated dense paths.

Trajectories span varied off-road terrains: (i) obstacle-rich (natural and artificial), (ii) extreme slopes (ditches, cliffs), and (iii) hybrid. We collect 15 teacher demos for (i) and (ii), and 20 for (iii). The evaluation set includes 8 trajectories for (i) and (ii), and 15 for (iii).

\begin{figure}[t]
    \centering
    \includegraphics[width=0.75\linewidth]{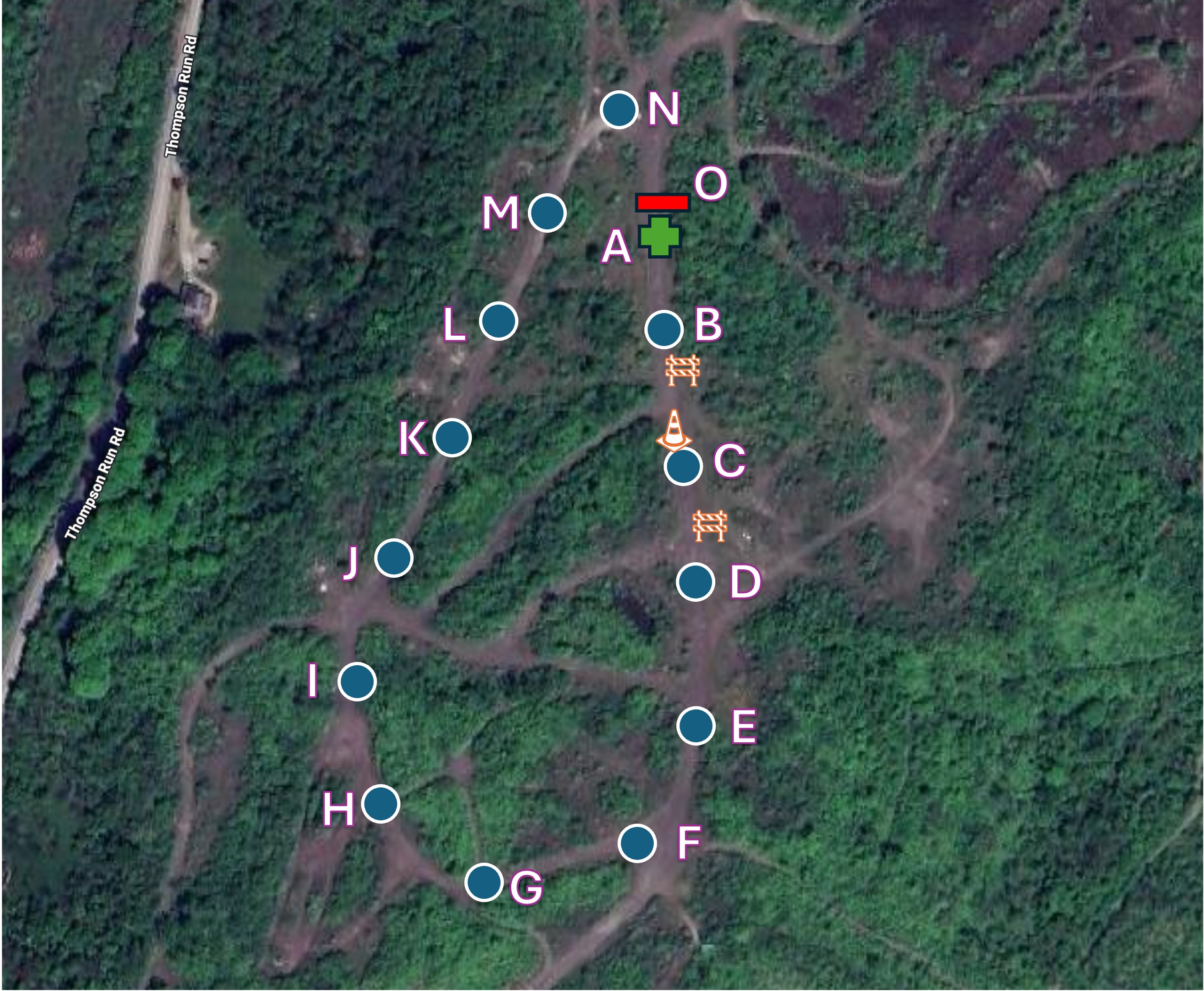}
    \caption{
    Waypoints supplied to the planning algorithm to generate the training trajectory.
    Groups of obstacles are placed along the trajectory, and the terrain is generally made up of uneven tracks.
    The test track is around 800m in length overall (A-O).
    We test the vehicle's dynamic handling over the entire course, but the obstacles are only placed randomly between B-D (around 120m).
    }
    \label{fig:eval_traj}
    \vspace{-8mm}
\end{figure}

\subsection{Simulation Evaluation Metrics}
\label{text:eval_metrics}
We evaluate policy performance across test trajectories, with the mode of the distribution as the selected action. All baselines utilize an A* planner to generate sparse waypoints. For each episode, we compute the following metrics:
\begin{itemize}
    \item Success Rate (\texttt{sr}): 1 if the vehicle reaches the goal within radius $r$, 0 otherwise.
    \item Completion Percentage (\texttt{cp}): Maximum progress toward the goal, normalized by the initial distance.
    \item Mean Speed (\texttt{ms}): Average vehicle speed during the episode.
\end{itemize}

\begin{table*}[th!]
    \centering
    \resizebox{\textwidth}{!}{%
    \begin{tabular}{l l ccc ccc ccc c}
    \toprule
      & Controller & \multicolumn{3}{c}{Extreme Slopes}  & \multicolumn{3}{c}{Obstacles} & \multicolumn{3}{c}{Hybrid} \cr
      \midrule
         & & \texttt{sr} & \texttt{cp} & \texttt{ms}  & \texttt{sr} & \texttt{cp} &  \texttt{ms}  & \texttt{sr} &  \texttt{cp}  & \texttt{ms} & \texttt{ti} \\
        \cmidrule(lr){3-5} \cmidrule(lr){6-8} \cmidrule(lr){9-11}
        &
        MPPI + Teacher
            & 0.88   &0.96   &5.83
            & 1.00   &1.00   &5.91
            & 0.94   &0.96   &5.69
            & 2.02
            \\
        \midrule
        \midrule
        \multirow{3}{*}{
        \shortstack[l]{MPC\\(Nonreal-time)}
        } &
        CEM\cite{algo:cem} + PID
            & 0.88 & 0.96 & 5.51
            & 1.00 & 1.00 & 5.16
            & 0.87 & 0.94 & 5.13
            & 3.47
            \\
        &
        MPPI\cite{algo:mppi} + PID
            & 0.88 & 0.96 & 5.39
            & 1.00 & 1.00 & 5.87
            & 0.87 & 0.94 & 5.43
            & 2.02
            \\
        &
        RL+MPPI\cite{mppi:rl+mppi} + PID
            & 0.88 & 0.96 & 5.26
            & 1.00 & 1.00 & 5.88
            & 0.87 & 0.94 & 5.40
            & 1.77
            \\
        \midrule
        \multirow{4}{*}{
        \shortstack[l]{MPC\\(Real-time)}
        } &
        CEM\cite{algo:cem} + PID
            & 0.38 & 0.49  & \textbf{5.52}
            & 0.25 & 0.38 & 5.16
            & 0.27 & 0.43 & 5.13
            & 0.13
            \\
        &
        MPPI\cite{algo:mppi} + PID
            & 0.38 & 0.57 & 5.43
            & 0.25 & 0.48 & \textbf{5.48}
            & 0.27 & 0.46 & 5.54
            & 0.12
            \\
        &
        RL+MPPI\cite{mppi:rl+mppi} + PID
            & 0.38 & 0.61 & 5.32
            & 0.25 & 0.50 & 5.46
            & 0.27 & 0.52 & \textbf{5.63}
            & 0.12
            \\
        &
        TADPO\rlap{$^{\dag}$}
        &\textbf{0.75}   &\textbf{0.87}   &4.99
        &\textbf{0.85}   &\textbf{0.96}   &5.26
        &\textbf{0.67}   &\textbf{0.88}   &5.30
        &\textbf{0.002}
        \\
      \midrule
      \midrule
      \multirow{7}{*}{\shortstack[l]{RL/IL\\(Real-time)}} &
            DAgger\cite{algo:dagger}
            &0.00   &0.58      &1.96
            &0.00   &0.83      &1.62
            &0.00   &0.79      &1.68
            & 0.002
            \\
            &
            PPO\cite{algo:ppo}
            &0.00   &0.14      &0.38
            &0.00   &0.25      &0.49
            &0.00   &0.37      &0.40
            & 0.002
            \\
            &
            PPO+BC
            &0.00   &0.25     &0.94
            &0.00   &0.40     &0.78
            &0.00   &0.32     &0.84
            & 0.002
            \\
            &
            SAC\cite{algo:sac}
            &0.00   &0.01     &1.71
            &0.00   &0.16     &1.64
            &0.00   &0.24     &1.61
            & 0.002
            \\
            &
            SAC+Teacher
            &0.00   &0.50     &1.21
            &0.00   &0.29     &1.24
            &0.00   &0.58     &1.24
            & 0.002
            \\
            &
            IQL\cite{algo:iql}
            &0.25   &0.49    &4.85
            &0.13   &0.71    &5.01
            &0.07   &0.76    &5.03
            & 0.002
            \\
            &
            TADPO\rlap{$^{\dag}$}
            &\textbf{0.75}   &\textbf{0.87}    &\textbf{4.99}
            &\textbf{0.85}   &\textbf{0.96}    &\textbf{5.26}
            &\textbf{0.67}   &\textbf{0.88}    &\textbf{5.30}
            & 0.002
            \\
        \bottomrule
    \end{tabular}
    }
    \caption{
    Our method ($^{\dag}$) compared with baselines, where \texttt{sr} denotes Success Rate, \texttt{cp} denotes Completion Percentage, \texttt{ms} denotes Mean Speed, and \texttt{ti} is the Time of Inference for one control step.
    (Real-time) denotes allotting a limited compute budget for the main control loop necessary for real-time deployment.
    MPPI+Teacher is used to provide supervision or demonstrations for RL and IL methods that require it.
    For RL and IL methods, the model architecture is kept constant to ensure fair comparison.
    }
    \label{tab:control_baselines}
    \vspace{-7mm}
\end{table*}

\begin{table}[b]
    \begin{tabular}{l ccc}
    \toprule
    Configuration  & \texttt{cte (m)} & \texttt{cp} & \texttt{ms (m/s)}\\
    \midrule
    Long distance High-speed Control     & 0.45  & 1.00   & 3.41 \\
    Obstacle Avoidance                   & 1.50   & 0.71   & 2.29 \\
    \bottomrule
    \end{tabular}
    \caption{
    Our method performance using three metrics: mean cross-track error (\texttt{cte}), completion percentage (\texttt{cp}), and mean speed (\texttt{ms}).
    The test track for the long-distance control configuration has a total length of 800\,m. Obstacle-avoidance performance is evaluated over 12 runs on a 120\,m track, with randomized barrel placements.
    }
    \label{tab:real_results}
\end{table}

\section{Real-World Vehicle Evaluation}
\subsection{Platform}
We deploy our policy on a Sabercat (pictured on the right in Figure \ref{fig:sim-real}), a 2-ton full-scale off-road vehicle designed to handle challenging terrain and equipped with skid steer control, forward-facing RGB camera, additional stereo cameras, and odometry sensors.
Its large size, high cost, and real-world dynamics make it a challenging platform for autonomous navigation, where any collision would result in significant damage.
Deploying RL policies on Sabercat allows us to evaluate vision-based obstacle avoidance and traversability reasoning under realistic conditions, including uneven terrain, variable surface types, and dynamic environmental changes, providing a compelling demonstration of zero-shot sim-to-real performance.
\subsection{Adaptations in Training Procedure and Observations Space}
In an effort to minimize the Sim2Real gap, we attempt to mimic a similar sensor setup and observation space from our simulator on the real vehicle with a few modifications.
Due to the restricted sensor configuration on the real vehicle, a detailed BEV local map is costly to construct with stereo depth.
Accordingly, we remove the top down camera from the observation space and decrease the waypoint spacing to $20$m and $4$m apart for the teacher and student respectively.

As a result, we train our \method{} Policy to output forward velocity and yaw rate given waypoint information and forward-facing image with the same rewards as above.
This controller is then deployed on the real vehicle with zero finetuning on real world images or scenarios.

More specifically, for vision features, we follow \cite{costmap:SALON,cv_offroad:traversibility} and use a frozen DinoV2 ViT-S/14\cite{vfm:dinov2} as our vision backbone to feed image features to our policy.
Additionally, we employ a Probabilistic Road Map (PRM) global planner from Open Motion Planning Library\cite{planning:ompl} to generate sparse waypoints.

Further, we train our \method{} controller using BeamNG\cite{sw:beamng} in an off-road forest environment with similar traffic barrels, and then deploy the trained policy zero-shot in a similar off-road forest environment in Pittsburgh, PA.

\subsection{Evaluation Trajectory}
To demonstrate the performance of our system in the real world, we set up a test track with two different configurations near Pittsburgh, PA (as shown in Figure \ref{fig:eval_traj}).

One configuration is designed to test long-range, high-speed control and evaluates the vehicle's ability to handle complex terrain dynamics at speed.
This requires the vehicle to traverse steep terrain at the traction limit and maintain control in tight turns.

The other configuration is designed to test obstacle avoidance and assesses the control policy's ability to plan over a long horizon and avoid unmapped obstacles.
For obstacles, we selected groups of traffic barrels and placed them between sparse waypoints along the vehicle's path.

The first configuration runs from A to O in the figure and covers a distance of approximately 800\,m.
The second configuration runs from B to D in the figure and covers a distance of approximately 120\,m.
We tested the second configuration over 12 runs with randomized barrel placements.

\subsection{Real World Evaluation Metrics}
We use similar metrics to our simulation metrics except replacing success rate with cross track error.
\label{text:real_metrics}
\begin{itemize}
    \item Cross Track Error (\texttt{cte}): Cross Track Error is the perpendicular distance from the linear interpolated path between dense waypoints to the vehicle center
    \item Completion Percentage (\texttt{cp}): Same as in Section \ref{text:eval_metrics}.
    \item Mean Speed (m/s) (\texttt{ms}): Same as in Section \ref{text:eval_metrics}.
\end{itemize}

\section{Results and Discussion}
\label{Results}

\subsection{Simulation: MPC Baselines}

In Table \ref{tab:control_baselines}, MPC (Nonreal-time) baselines demonstrate the performance of CEM, MPPI and RL+MPPI controllers with a long planning horizon $h$ and high number of samples $N$.
These controllers calculate the next waypoints while the simulation is paused, enabling them to determine the next action before resuming the simulation.
The results show that, given a sufficient number of samples and an extended planning horizon, these controllers achieve similar levels of performance.
The cost function and dynamics model implementation are similar to that of \cite{mpc_offroad:aggressive_uw}.

MPC (Real-time) baselines demonstrate the performance of these controllers when operating under real-time constraints with a limited computational budget.
In contrast to MPPI, CEM employs a more iterative approach to sampling and evaluating action sequences, which results in a higher computational cost. RL+MPPI improves upon MPPI by incorporating a learned terminal value function and a state-dependent action distribution, reducing the $N$ and $h$ requirement.
However, all of these methods experience a substantial decline in performance when executed with a reduced computational budget for real-time inference.

\subsection{Simulation: RL and IL Baselines}

We evaluate \method{} against established RL and imitation learning techniques in Table \ref{tab:control_baselines}, adapted as necessary for our domain.
When applicable, all policies are supplied with the same teacher policy $\mu$ trained in  \ref{offroad_training}.
The model architecture is held constant across all the policies to ensure a fair comparison.
The result is shown in Table \ref{tab:control_baselines}.

\textbf{DAgger} is used to distill the behavior of the teacher through supervised learning.
In long-horizon planning tasks, DAgger suffers from compounding errors. As the policy accumulates errors and deviates from expert trajectories, it encounters previously unseen states, leading to significant performance degradation compared to the teacher policy.

The \textbf{PPO} policy is trained analogously to the teacher policy $\mu$ described in Section \ref{offroad_training}, but using only sparse waypoints.
As explained in Section \ref{ppo_explain}, PPO faces challenges in effective exploration and struggles to differentiate between various terrain types and obstacles, resulting in a suboptimal, overly cautious navigation strategy.

The \textbf{PPO + BC} policy aims to distill teacher behavior by adding a KL divergence loss to the PPO loss function as $L^\text{KL} = L^{\text{PPO}} - \beta \text{KL}[\pi(a_t | s_t^\pi), \mu(a_t | s_t^\mu)]$, introducing a term that aligns the policy $\pi$ with the teacher policy $\mu$ across all encountered states.
While this method provides strong supervision, it encounters challenges similar to DAgger when the student queries the expert from out-of-distribution states. Moreover, the unconstrained updates from the KL divergence term lead to training instability, resulting in convergence to a suboptimal policy.

The \textbf{SAC} policy struggles due to entropy maximization, which leads to excessive exploration of irrelevant states and reduced focus on task-specific goals, making it less effective in environments requiring targeted exploration across multiple distinct tasks.

The \textbf{SAC + Teacher} policy incorporates teacher demonstrations into the SAC framework by pre-populating a portion of the replay buffer.
We maintain consistency with \method{} by using an equivalent buffer size and setting the teacher trajectory ratio to $p=0.5$.
However, SAC's performance degrades in multi-task problems \cite{sac_multitask}.

The \textbf{IQL} policy is trained using teacher demonstrations to reinforce behavior by learning the Q-values associated with those actions.
Although IQL demonstrates some success in navigating steep slopes, its overall performance lags behind \method{}, as it is known to encounter difficulties in multi-task problems, such as off-road autonomy \cite{iql_multitask}.

\paragraph*{\textbf{TADPO}}

The \method{} policy's success rate (\texttt{sr}) and completion percentage (\texttt{cp}) notably exceed those of other real-time baseline methods.
Furthermore, the policy achieves a comparably high mean speed (\texttt{ms}) across all test trajectory sets.
Ablation studies show that $\epsilon_\mu=0.5$ and constant $p=0.5$ yield the best performance, used in baseline comparisons.

\subsection{Real-World Evaluation}
We demonstrate and evaluate the trained student policy on the full-scale Sabercat vehicle for both long-distance high-speed control and obstacle avoidance as shown in Figure \ref{fig:sabercat}.
As seen in Table \ref{tab:real_results}, a policy trained using \method{} achieves a high percentage of obstacle avoidance and minimum cross-track error for long-distance high speed control demonstrating robust navigation and waypoint tracking without any fine-tuning on the real vehicle. A higher cross-track error when encountering obstacles occurs because the vehicle deviates from the desired path to avoid them and then returns to the path afterward. The policy modulates its speed to safely navigate around obstacles.

\section{CONCLUSIONS}

We introduce \method{}, an extension of PPO that enables simultaneous learning from expert demonstrations and on-policy environment interactions to tackle long-horizon planning and hard exploration challenges.
By training a \method-based policy, we develop an end-to-end off-road autonomy pipeline capable of real-time, long-range navigation in complex, obstacle-rich, and diverse terrains.
Our experiments show that \method{} outperforms RL and IL baselines, validating its effectiveness.
Our experiments demonstrate strong performance in simulation and, importantly, zero-shot sim-to-real transfer on a full-scale off-road vehicle. This work represents, to our knowledge, the first deployment of end-to-end RL-based policies on a full-scale off-road platform.
Our future work includes extending this framework to more diverse terrains.

%%%%%%%%%%%%%%%%%%%%%%%%%%%%%%%%%%%%%%%%%%%%%%%%%%%%%%%%%%%%%%%%%%%%%%%%%%%%%%%%

\section{ACKNOWLEDGMENT}
This work was supported in part by the U.S. Army Research Office and the U.S. Army Futures Command under Contract No. W519TC-23-C-0030. The authors would like to acknowledge the contributions of Prasanna Kannappan, Anoushka Alavilli, Sam Zieger, Jessica Kasemer, and Daniela Resasco. The authors would like to thank BeamNG GmbH for the academic license for BeamNG.tech.

\bibliographystyle{IEEEtran}
\bibliography{references}

\end{document}